# Associative Memory using Attribute-Specific Neuron Groups-2: Learning and Sequential Associative Recall between Cue Neurons for different Cue Balls

**Hiroshi Inazawa**

*Center for Education in Information Systems, Kobe Shoin Women's University[1],*
*1-2-1 Shinohara-Obanoyama, Nada Kobe 657-0015, Japan.*
**E-mail**: inazawa706@gmail.com

## Abstract

This paper introduces a neural network model that learns multiple attributes as images and performs associated, sequential recall of the learned memories. Briefly, the model presented here is an associative memory model that extends previous models [1] by increasing the number of attributes. In the real world, memory recall generates a chain of associations consisting of complex and diverse data with meaningful relations. However, because this experimental system is designed to implement and verify the processing operations behind such operations, we believe it is not a problem if the associative memory (i.e., the chain of data) is composed of attributes that do not necessarily have clear relation with each other. Accordingly, the attribute-processing systems prepared in this study consist of five types: the C.CB-RN system for processing color attributes, the S.CB-RN system for shape attributes, and the V.CB-RN system for size attributes, as adopted in our previous paper [1], as well as the SV.CB-RN system for processing the names of the world's most beautiful scenery (spectacular view names) and the CN.CB-RN system for processing constellation names. As before, the data presented to each CB-RN system are represented as image patterns using QR codes [2]. These five types of CB-RN systems will be combined and trained with QR code pattern images of the attribute elements of each system. After that, when a pattern image of an attribute element is presented to any of the CB-RN systems, a mechanism will be constructed in which a chain (associative) recall of pattern images of related attribute elements in the other trained systems will be generated.

## 1. Introduction

This paper introduces a neural-network-based associative memory model [3–9] that learns attributes as images and enables associated, sequential recall through the integration of multiple learned memories. In other words, we investigate an extended version of the previous attribute associative memory model [1] by increasing the number of attributes. The neural network system used in this study is the Cue Ball-Recall Net (CB-RN) system, which was employed in our previous work [1] [10] and is hereafter referred to as the CB-RN system, or simply CB-RN. We set the number of associated attributes to five, motivated by reports suggesting that humans can associate approximately five distinct memories simultaneously [31–34]. In this model, as in our previous work, attributes encode as characters and processed as pattern images of the QR-code [2]. In this study, the element names (character strings) constituting each attribute are represented using QR-code pattern images. As a result, the scope of applicability is greatly expanded, enabling relatively straightforward representation of subtle tastes, odors, and other sensory modalities. The CB-RN system first learns QR code pattern images that represent multiple attribute elements within each subsystem. After learning, when a single image is presented to an arbitrary system, the four other images that were memorized in association with it are sequentially recalled in their corresponding CB-RN systems. In the real world, memory recall generates a chain of associations consisting of complex and diverse data with meaningful relations. However, because this experimental system is designed to implement and verify the processing operations behind such operations, it is not a problem if the associative memory (i.e., the chain of data) is composed of attributes that do not necessarily have clear relation with each other. In this paper, five attributes were prepared: Color, Shape, Volume, Spectacular View and Constellation name. Accordingly, the CB-RN systems configured in this study comprise five types: the C.CB-RN system for processing Color, the S.CB-RN system for Shape, and the V.CB-RN system for Volume, which were adopted in our previous paper [1], as well as the SV.CB-RN system for processing the names of Spectacular View around the world and the CN.CB-RN system for processing Constellation names. While it is in principle possible to extend the model considered here to actual real-world objects, doing so would significantly increase both the number of attributes and the number of elements that compose them. For example, with respect to the color attribute, as many as 465 distinct color categories are used in Japan. The same holds true for shape. Therefore, applying this model directly to real-world objects from the outset is considered to be beyond the scope of this paper. The basic operation of the model follows the well-known dynamics of associative memory [3–9]. In associative memory, recalling a single item often triggers the successive recall of related memories, a phenomenon that is commonly observed in human memory retrieval. In particular, memory recall appears to be closely related to imagery, and we believe that the model presented here captures at least part of the underlying mechanisms.

---

[1] This was my affiliated institution until the end of March 2025, when I retired. Please note that as of April 2025, the university name has been changed from "Kobe Shoin Women's University" to "Kobe Shoin University."

In recent years, AI—particularly generative AI—has undergone remarkable advances and has entered a practical stage of deployment [11–19]. Broadly speaking, these approaches fall under the research field known as deep learning, which is based on multilayer neural network techniques developed during the 1980s and early 1990s. Meanwhile, memory-focused models appear to have received relatively little attention compared to the learning models used in generative AI. Memory models in neural networks are generally associative memory models, which, like human memory processing, can recall entire stored information from presented fragments and sequentially search for related information [3, 6–9, 28, 29]. We consider that humans probably learn by repeating stored memories in their brains. Therefore, we consider that the research on memory mechanisms are an extremely important field, alongside studies of learning mechanisms. It should be noted that, in recent years, the research of memory based on transformer theory has been explored [30]. In Section 2, the details of the model are described, Section 3 presents the simulation results and Section 4 provides conclusions and discussion.

## 2. Specification of the model

In this section, we describe the details of the model. Although much of this description overlaps with that presented in Ref. [1], we have written them accurate to preserve the independence of this paper. The proposed model consists of a system (hereinafter referred to as the CB-RN system) consisting of a cluster of neurons (hereinafter referred to as the Cue Ball) containing a large number of neurons (hereinafter referred to as the cue neurons) and a two-dimensional spatially extended screen called a recall net that represents the image. In addition, the number of neurons in the recall net (hereafter referred to as recall neurons) is determined by the number of pixels in the image. Note that in this study, all images are assumed to be of the same size, and therefore the number of recall neurons in the Recall Net is also kept constant. In reality, one Recall Net would be sufficient. However, to make the explanation and image of the model easier to understand, we decided to place one recall net for each Cue Ball. Learning occurs between cue neurons and recall neurons, and between cue neurons within different Cue Balls. One final thing to note is that one feature of this model is that one cue neuron memorizes[2] the entire image, so as the number of memories stored increases, so does the number of cue neurons.

As described in the Introduction, this paper considers five types of CB-RN systems, each corresponding to a specific attribute: C.CB-RN for Color processing, S.CB-RN for Shape processing, V.CB-RN for Volume processing, SV.CB-RN for Spectacular View processing, and CN.CB-RN for Constellation processing. Figure.1 illustrates the conceptual structure of these CB-RN systems.

---

[2] The grandmother cell hypothesis has been proposed, in which a single neuron is responsible for memory.

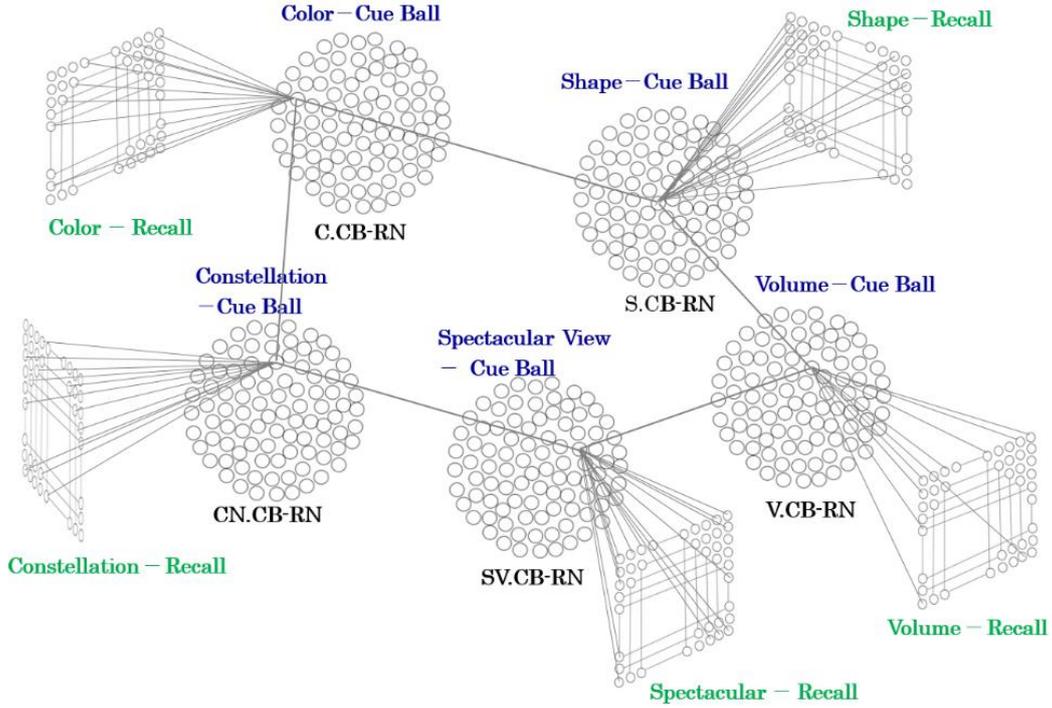

Figure.1: Schematic diagram of the CB-RN system. Each Cue Ball is represented as a sphere, and one Recall Net is assigned to each Cue Ball. All recall neurons in each Recall Net are connected to all cue neurons in the corresponding Cue Ball. There are no connections between recall neurons, nor between cue neurons of the same cue ball. However, there are connections between cue neurons of different cue balls. In principle, cue neurons of different cue balls can be connected freely. However, for convenience of simulation processing, this paper is that the connections between cue neurons of different cue balls only exist in the following order: Color−Cue Ball ⇔ Shape−Cue Ball ⇔ Volume−Cue Ball ⇔ Spectacular View−Cue Ball ⇔ Constellation−Cue Ball.

In this paper, each attribute element within a Cue Ball is represented as a character and displayed on the Recall Net as a QR-code pattern image. For use in the simulation, the pixel values of the generated QR code pattern image are converted into digital values and input to each neuron. Note that each QR-code pattern image consists of 116 × 116 pixels and is converted into a 13,456-dimensional vector of digital values for use in the simulations.

Following, we explain the basic skeleton of the mechanism: a cue neuron in the cue ball is connected to every recall neuron in the corresponding recall net. There are no internal connections among recall neurons within a Recall Net. Cue neurons are connected to all cue neurons in different Cue Balls. Note that there are no connections among cue neurons within the same Cue Ball. In this study, we assumed a simulation process and set the connections between cue neurons to exist only in the following order: "Color-cue ball ⇔ Shape-cue ball ⇔ Volume-cue ball ⇔ Spectacular scenery-cue ball ⇔ Constellations-cue ball." Each cue neuron has a threshold and outputs a normalized value of "0.0" or "1.0" based on the intermediate value. In principle, the number of cue neurons in a cue ball can be increased without limit, but in this study, the number of cue neurons in each cue ball was fixed at the number of attribute elements in the prepared dataset, i.e., seven types. Because all QR code image patterns were sized to the same resolution, the number of recall neurons (i.e., vector dimension) in each recall net is 116 × 116, resulting in a 13,456-dimensional vector. Next, we explain the specific processing steps. In the following, first, we explain the learning process, and then the recall process.

An overview of the learning (encoding) operation is as follows. First, a single QR-code pattern image to be learned (represented as digital values) is presented to one Recall Net. Subsequently, learning is performed between all recall neurons within this Recall Net and a single cue neuron (within the corresponding Cue Ball) to which they are connected. Specifically, after the recall neuron processes the input from this cue neuron, $w_{ji}^{attr}$ learns using the output data of the recall neuron and the image pattern data of the presented QR code. In this case, learning corresponds to the adjustment of the synaptic weights of the recall neurons, denoted as $w_{ji}^{attr}$. The superscript "$attr$" is an identifier used to distinguish each Cue Ball and represents a specific attribute (e.g., "Color"). After this, the outputs of all recall neurons in this recall net are fed back to the cue neurons trained by these recall neurons, and the cue neurons are trained by adjusting their synaptic weights (denoted as $v_{ij}^{attr}$). Through these learning processes, bidirectional learning between the cue neuron and all recall neurons within the Recall Net is completed. These processes are performed for five CB-RN systems corresponding all attributes (Color, Shape, Volume, Spectacular View, and Constellation). After these steps are completed, the next learning step involves learning between cue neurons belonging to different cue balls. For example, learning is performed between the $k$-th cue neuron in the Color−Cue Ball and the $l$th cue neuron in the Shape−Cue Ball, and so on.

Next, we provide an overview of the recall process for learned images. To check whether the synaptic weights of the recall neurons, $w_{ji}^{attr}$, have been learned correctly, the outputs of the learned cue neurons in the Cue Ball are input to all corresponding recall neurons in the recall net, and the resulting QR code image pattern displayed in the recall net is checked. Also, the success of learning of the cue-neuron synaptic weights, $v_{ij}^{attr}$, can be verified through the following procedure. The learned QR code pattern image is presented to a recall net, and the output of every recall neuron in this recall net is input to every cue neuron in the corresponding Cue Ball. At this stage, each cue neuron generates an output. If a cue neuron whose output value is close to the learning target value $\theta$ specified during training is found, it is determined that the cue neuron has learned the corresponding pattern. For verification, the output of this cue neuron is normalized to 1.0 using a threshold function and fed back to all recall neurons in the corresponding recall net, resulting in the recall of the learned (presented) pattern. Note that the threshold value of the threshold function is determined by the simulation process. Finally, we describe the recall process between cue neurons belonging to different Cue Balls. When the learned QR code image pattern is presented to the recall net, the corresponding learned cue neuron (e.g., the $k$-th neuron in Color−Cue Ball) in the associated Cue Ball output 1.0. This output is provided as input to all cue neurons in another scheduled Cue Ball. In this Cue Ball, only the trained cue neuron (e.g., the $l$-th neuron in Style−Cue Ball) produces an output of 1.0, indicating that learning has been successfully completed.

Now, let us describe these learning and recall processes more explicitly using mathematical formulations. We first describe the learning process, followed by the recall process. In the recall neuron learning process ($w_{ji}^{attr}$ learning), input is made from a single cue neuron of any cue ball to all recall neurons in the corresponding recall net. The input-output relation (the output of the recall neuron, $y_j^{attr}$) at this time is expressed as follows:

$$y_j^{attr} = w_{ji}^{attr} x_i^{attr} \quad , \tag{1}$$

where $x_i^{attr}$ denotes the output of the $i$-th cue neuron, and $y_j^{attr}$ denotes the output of the j-th recall neuron. $w_{ji}^{attr}$ denotes the synaptic weight from the $i$-th cue neuron to the $j$-th recall neuron in the corresponding Recall Net. Although Eq. (1) is a commonly used formulation, it is important to note that no summation is taken over the index $i$. Here, as mentioned above, "$attr$" indicates the attribute (Color、Shape、Volume、Spectacular View、Constellation) used to distinguish each Cue Ball. Table 1 also summarizes the attribute identifiers and components of each attribute for each Cue Ball.

$attr \equiv a \sim e = C(= Color), S(= Style), V(= Volume), SV(= Spectcular\ View), CN(= Constelation)$

| C | S | V | SV | CN |
|---|---|---|---|---|
| 0. $red$ | 0. $square$ | 0. $extra-large$ | 0. $Iguazu$ | 0. $Andromeda$ |
| 1. $orange$ | 1. $circle$ | 1. $large$ | 1. $MaunaKea$ | 1. $Aquarius$ |
| 2. $yellow$ | 2. $oval$ | 2. $medium$ | 2. $MilfordSound$ | 2. $Cassiopeia$ |
| 3. $green$ | 3. $rectangle$ | 3. $small-medium$ | 3. $MonumentVY$ | 3. $Centaurus$ |
| 4. $blue$ | 4. $trapeziod$ | 4. $small$ | 4. $RockiesMT.$ | 4. $Cygnus$ |
| 5. $indigo$ | 5. $triangle$ | 5. $extra-small$ | 5. $eTkapo$ | 5. $Orion$ |
| 6. $purple$ | 6. $rhombus$ | 6. $mini$ | 6. $Yellowknife$ | 6. $Perseus$ |

Table.1: Cue Ball Attribute Identification and Attribute Elements. Each cue ball's identification code is displayed at the top, with its attribute elements numbered 0 to 6 below. In actual use, only the attribute element name (such as the string "$red$" without a leading index) is converted into a QR code image pattern for use in the model.

Also, when converting to QR codes, the textual labels describing each attribute element (e.g., "$red$") are transformed into QR-code image patterns; the leading numerical indices shown in the table are not included in the images. Learning is performed using the gradient descent method (GDM) [22, 23], so the error function $E^{attr}$ in this case is expressed as follows:

$$E^{attr} \equiv \frac{1}{2}\sum_{j=0}^{M}(d_j^{attr\ p} - y_j^{attr})^2 \quad, \tag{2}$$

where $d_j^{attr\ p}$ denotes the element value of the QR-code pattern presented to the Recall Net and "$p$" represents the pattern number (which also corresponds to the attribute element number), and $M$ represents the number of recall neurons in one recall net ($M = 116 \times 116 - 1 = 13{,}455$). The weight update rule derived using the GDM is given as follows (see [1] for details).

$$w_{ji}^{attr}(t+1) = w_{ji}^{attr}(t) + \Delta w_{ji}^{attr}(t) \tag{3}$$

$$\Delta w_{ji}^{attr}(t) = -\varepsilon_W \frac{\partial E^{attr}}{\partial w_{ji}^{attr}} = \varepsilon_W \left(d_j^{attr\ p} - y_j^{attr}(t)\right) x_i^{attr}(t) \quad, \tag{4}$$

where $t$ is the number of updates and $\varepsilon_W$ is the learning rate, which is common to all recall nets. The learning rate is set to $\varepsilon_W = 1.0$. Using the input–output relation in eq. (1) together with the weight learning rules in Eqs. (3) and (4), we can show that $w_{ji}^{attr}$ can reliably learn the presented patterns.

$$\begin{aligned} y_j^{attr}(t+1) &= w_{ji}^{attr}(t+1)x_i^{attr}(t+1) \\ &= (w_{ji}^{attr}(t) + (d_j^{attr\ p} - w_{ji}^{attr}x_i^{attr}(t))x_i^{attr}(t+1) \end{aligned} \tag{5}$$

Assuming there is an output from the $i$-th cue neuron, and setting "$x_i^{attr}(t) = x_i^{attr}(t+1) = 1.0$", eq. (5) becomes as follows.

$$y_j^{attr}(t+1) = d_j^{attr\ p} \tag{6}$$

In this way, after training, "$y_j^{attr}(t+1)$" matches the image pattern (i.e., attribute elements) of the presented QR code.

Next, let us look at the learning process of the synaptic weights from the recall neuron to the cue neuron. The input-output relation from the recall neuron to the cue neuron (the output of the cue neuron, $x_i^{attr}$) is given as follows:

$$x_i^{attr} = f(q_i^{attr}) = f\left(\sum_{j=0}^{M} v_{ij}^{attr} y_j^{attr}\right) = \begin{cases} 0 & for\ q_i^{attr} < D \\ 1 & for\ q_i^{attr} \geq D \end{cases}, \quad (7)$$

where $q_i^{attr}$ denotes the intermediate output value of the $i$-th cue neuron before thresholding, and $y_j^{attr}$ represents the input value from the $j$-th recall neuron. In this case, we use the value $y_j^{attr}$ that is obtained from "$x_i^{attr} = 1.0$" using learned $w_{ji}^{attr}$. $v_{ij}^{attr}$ denotes the synaptic weight from the j-th recall neuron to the $i$-th cue neuron, where $f$ is a threshold function, and $D$ is a threshold value common to all cue neurons of each Cue Ball. The memory (memory) training is performed using the same GDM as in the training of $w_{ji}^{attr}$. In this case, the error function is defined as follows:

$$e^{attr} \equiv \frac{1}{2} \sum_{i=0}^{L} (\theta - q_i^{attr})^2 , \quad (8)$$

where $\theta$ is an arbitrary (constant) value specified as the learned value of $v_{ij}^{attr}$, and it is set identically for all cue neurons within each Cue Ball. In addition, $L$ is equal to the number of attribute elements within each Cue Ball minus one, and in the present setting "$L = 6$." Using these definitions, the weight update rule derived by GDM is given as follows.

$$v_{ij}^{attr}(t' + 1) = v_{ij}^{attr}(t') + \Delta v_{ij}^{attr}(t') , \quad (9)$$

$$\Delta v_{ij}^{attr}(t') = -\varepsilon_V \frac{\partial e^{attr}}{\partial v_{ij}^{attr}} = \varepsilon_V \left(\theta - q_i^{attr}(t')\right) y_j^{attr}(t') , \quad (10)$$

where $t'$ denotes the update iteration index, and $\varepsilon_V$ is the learning rate. We use the input value $y_j^{attr}$ that is obtained from Eq. (1) after the learning of $w_{ji}^{attr}$, in the same manner as described above. Note that the learning rate $\varepsilon_V$, like $\varepsilon_W$, is set to 1.0 for all cue neurons of each Cue Ball. Similar to the learning of $w_{ji}^{attr}$, by using the input-output relation in eq. (7), the synaptic weight learning rules in Eqs. (9) and (10), and the conditions for $y_j^{attr}$ set below, we can reliably learn the weight $v_{ij}^{attr}$ as shown below. Note that in this case, $x_i^{attr}$ is assumed to have an output value of 1.0 from eq. (7) as a result of the input from the Recall Net.

$$q_i^{attr}(t' + 1) = \sum_{j=0}^{M} v_{ij}^{attr}(t' + 1) y_j^{attr}$$
$$= \sum_{j=0}^{M} v_{ij}^{attr}(t') y_j^{attr} \left(1 - \sum_{k=0}^{M} (y_k^{attr})^2\right) + \theta \sum_{j=0}^{M} (y_j^{attr})^2 , \quad (11)$$

where we consider that $y_j^{attr}$ in eq. (1) is equal to $d_j^{attr\ p}$, and impose the following normalization condition on $y_j^{attr}$.

$$\sum_{j=0}^{M}(y_j^{attr})^2 = \sum_{j=0}^{M}(d_j^{attr\ p})^2 \equiv 1 \quad (12)$$

By applying eq. (12), eq. (11) can be rewritten as follows.

$$q_i^{attr}(t'+1) = \theta \quad (13)$$

Thus, after learning $v_{ij}^{attr}$, $q_i^{attr}(t'+1)$ coincides with the learned value $\theta$. Applying the threshold function in eq. (7) to $q_i^{attr}(t'+1)$ in eq. (13) yields "$x_i^{attr}(t'+1) = 1.0$."

Finally, we will discuss learning between cue neurons in the Cue Ball. In actual processing, attribute chains are set as "$a \Rightarrow b \Rightarrow c \Rightarrow d \Rightarrow e$" and "$e \Rightarrow d \Rightarrow c \Rightarrow b \Rightarrow a$" (see Table 1). To simplify the explanation, we will use only the attribute chain "$a \Rightarrow b$" as an example below. Note that during these learning processes, the cue neurons connecting each Cue Ball and its corresponding Recall Net are assumed to have been already learned. The output of the kth cue neuron for the $a$ −Cue Ball (e.g., $a$ represents "$Color$"), "$x_k^a = 1.0$", which responded to the pattern presented to the recall net, is used to learn the $l$-th cue neuron associated with the $b$ −Cue Ball (e.g., $b$ represents "$Shape$"). The input-output relation between cue neurons is expressed as follows in the same format as eq. (7).

$$x_l^b = f(q_l^b) = f(u_{lk}^b x_k^a) = \begin{cases} 0 & for\ q_l^b < D \\ 1 & for\ q_l^b \geq D \end{cases}, \quad (14)$$

$u_{lk}^b$ denotes the connection weight from the $k$-th cue neuron of $a$ −Cue Ball to the $l$-th cue neuron of $b$ −Cue Ball. This means that each cue neuron has two types of connection weights: $v_{lk}^{attr}$, which receives input from the recall net, and $u_{lk}^{attr}$, which receives input from the other Cue Balls. $f$ is the threshold function and $D$ is the common threshold applied to all cue neurons in each Cue Ball. A schematic diagram of the input–output relations between these Cue Balls is shown in figure.2.

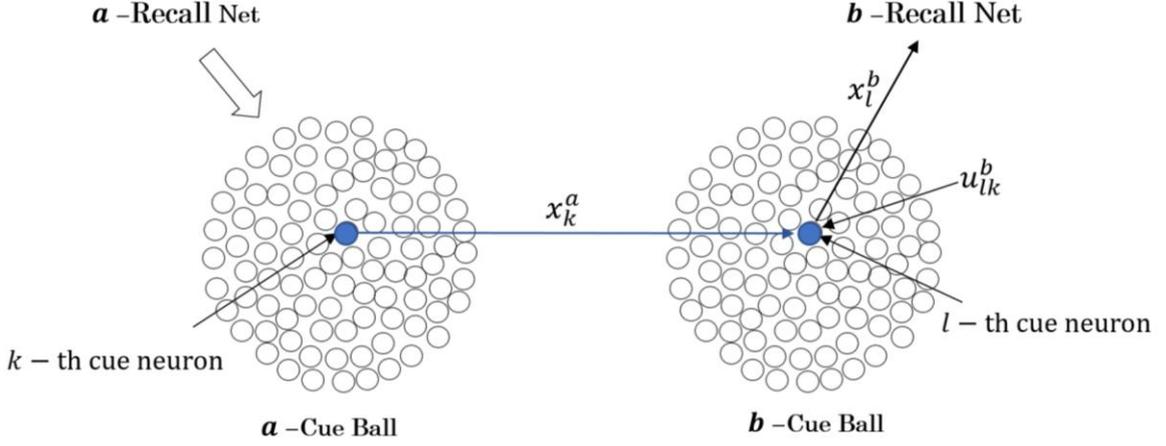

Figure 2: Variables between cue neurons for different Cue Balls. The input and output variables and connection weights of the cue neuron between two Cue Balls ($a \sim e$ =Color, Shape, Volume, Spectacular View, Constellation) have been are shown. When an image pattern is presented to $a$ −Recall Net, the kth cue neuron in $a$ −Cue Ball outputs $x_k^a$, which in turn outputs the output $x_l^b$ from the learned cue neuron in $b$ −Cue Ball. As a result, the learned QR-code pattern image is displayed in $b$ −Recall Net.

Next, we show the learning equation for the cue neuron of $b$ −Cue Ball. First, we define the error function of $b$ −Cue Ball as follows:

$$\eta^b \equiv \frac{1}{2}\sum_{l=0}^{L}(\theta - q_l^b)^2 \quad , \tag{15}$$

$\theta$ is the same as the parameter defined as the learning value of $v_{ij}^{attr}$ in eq. (8). Here, it is assigned as the learning value of $u_{lk}^b$ and is treated as the same constant value for all Cue Balls. Using these definitions, the weight update equation derived by the GDM is given as follows.

$$u_{lk}^b(t+1) = u_{lk}^b(t) + \Delta u_{lk}^b(t) \quad , \tag{16}$$

$$\Delta u_{lk}^b(t) = -\lambda_{CB}\frac{\partial \eta^b}{\partial u_{lk}^b} = \lambda_{CB}\left(\theta - q_l^b(t)\right)x_k^a(t) \quad , \tag{17}$$

where $t$ is the update iteration number and $\lambda_{CB}$ is the learning rate. Like $\varepsilon_W$ and $\varepsilon_V$, $\lambda_{CB}$ is set to 1.0 for cue neurons in all Cue Ball. As in the previous cases, by using the input–output relation in eq. (14) and the weight learning equations in Eqs. (16) and (17), the connection weights $u_{lk}^b$ can be reliably learned. In this case, the learned pattern image is presented to Recall Net a, so the output of the kth cue neuron of $a$ −Cue Ball is "$x_k^a$=1.0". Using this, the following equation can be obtained from Eqs. (14), (16), and (17):

$$q_l^b(t+1) = u_{lk}^b(t+1)x_k^a$$

$$= u_{lk}^b(t) + \theta - q_l^b(t) = \theta \tag{18}$$

Therefore, after learning $u_{lk}^b$, $q_l^b(t+1)$ will match the learned value $\theta$. Note that applying the threshold function in eq. (14) to $q_l^b(t+1)$ in eq. (18) yields "$x_l^b(t+1) = 1.0$."

## 3. Simulation and Results

In this section, we describe simulations of the learning and recall processes introduced in Section 2. The pattern data used in the simulation is obtained by expressing the attribute elements as characters and we use converting them into QR code image patterns, as explained in Section 2. As described above, each QR-code pattern image consists of 116 × 116 = 13,456 pixels, which are converted into a 13,456-dimensional vector of digital values for use in the simulations. There are five CB-RN systems used: C.CB-RN, S.CB-RN, V.CB-RN, SV.CB-RN, and CN.CB-RN. In the learning process, processing between the Cue Ball and the Recall Net is performed first. In this procedure, the attribute element images set for each cue ball are extracted from the top down (see Table .1), and the connection weights $w_{ji}^a$ and $v_{ij}^a$ are learned. Subsequently, learning is performed between cue neurons of different Cue Balls. In this learning procedure, the process is first executed in the order "$a \Rightarrow b \Rightarrow c \Rightarrow d \Rightarrow e$" (denoted by cmb = 0), and then in the order "$e \Rightarrow d \Rightarrow c \Rightarrow b \Rightarrow a$" (denoted by cmb = 1), where see Table 1. First, we present the Recall Net with an image pattern of an arbitrary QR code that has already been learned. In this state, the cue neuron that is learning the presentation pattern is outputting (e.g., $x_k^{Color} = 1.0$). Using this output, the next Cue Ball is learned by the cue neuron using equations (14), (16), and (17). Furthermore, the output of this second cue neuron (e.g., $x_l^{Style} = 1.0$) is provided as the input to the cue neuron of the subsequent Cue Ball, and the same learning procedure is applied, until the final Constellation Cue Ball. Note that, corresponding to cmb=0 and 1, the chains of two cue neuron numbers are learned in sequence. If cmb=0, then in the first chain (sequence 1), the cue neurons are learned in the order "0th⇒1st⇒2nd⇒3rd⇒4th", but in the second chain (sequence 2), the cue neurons are learned in the order "0th⇒4th⇒3rd⇒2nd⇒1st." Next, if cmb=1, then in the first chain (sequence 1), the cue neurons are learned in the order "0th⇒6th⇒5th⇒4th⇒3rd", but in the second chain (sequence 2), the cue neurons are learned in the order "0th⇒3rd⇒4th⇒5th⇒6th." The cue neuron indices (e.g., 0th, 1st, etc.) correspond to the respective attribute element indices (see Table 1). In the first sequence of each, the learning coefficient θ is set to 100, but in the second sequence it is set to θ = 110, producing in the results in a different learning process.

Below, we describe the simulation algorithms for these learning and recall processes. First, we describe the learning of $w_{ji}^{attr}$.

[Algorithm 1 — Learning of $w_{ji}^{attr}$]

1. Specify any Cue Ball.
2. Initialize each input variables, coupling coefficients, and the learning rate:
   $x_i^{attr} = 1.0$, $w_{ji}^{attr} = 0.0$, $\varepsilon_W = 1.0$
3. The element values of each image pattern presented to the Recall Net are substituted into Eqs. (4), (5), and the synaptic weights are learned.
4. Save the learned values of $w_{ji}^{attr}$ to a file.
5. Repeat processes 1–4 while varying the Cue Ball.

Next, the learning algorithm for $v_{ij}^{attr}$ is written as follows:

[Algorithm 2 — Learning of $v_{ij}^{attr}$]

1. Specify any Cue Ball.
2. Initialize each input and output variable, connection coefficients, learning rate, and set learning values and thresholds:
   $x_i^{attr} = 1.0$, $y_j^{attr} = 1.0$, $v_{ij}^{attr} = 0.0$, $\varepsilon_V = 1.0$, $\theta = 100.0$, $D = 72.0$
   ※ The threshold D = 72.0 has been set based on the simulation results.
3. Using the cue neuron output $x_i^{attr}$ obtained in equation (7), $y_j^{attr}$ is calculated from equation (1).
4. For all cue neurons of the specified cue ball, calculate $x_i^{attr}$ (and $q_i^{attr}$) in the same way as in step 3.
5. Using the $y_j^{attr}$ and $q_i^{attr}$ calculated in step 4, we learn the connection weights defined in equations (9) and (10).
6. Save the learned values of $v_{ij}^{attr}$ to a file.
7. Repeat steps 2–6 while varying the Cue Ball.

Finally, we describe the algorithm for the learning process between cue neurons of different Cue Balls.

[Algorithm 3 — Learning of $u_{kl}^{attr}$]

1. Define two processing groups for the Cue Balls, denoted by cmb = 0,1:
   cmb = 0 [a = Color → b = Shape → c = Volume → d = Spectacular View → e = Constellation]
   cmb = 1 [a = Constellation → b = Spectacular View → c = Volume → d = Shape → e = Color]
2. Initialize each input variables, coupling coefficients, and the learning rate and setting the learned parameters and thresholds:
   $x_k^{attr} = 1.0$, $u_{kl}^{attr} = 0.0$, $\lambda_{CB} = 1.0$, $\theta = 100.0$, $D = 72.0$

3. For cmb=0,1, processing is performed in the order of the cue neuron numbers (image patterns) of the first and second series defined below.:

   cmb = 0: 1st series [0th ⇒ 1st ⇒ 2nd ⇒ 3rd ⇒ 4th]   2nd series [0th ⇒ 4th ⇒ 3rd ⇒ 2nd ⇒ 1st]
   cmb = 1: 1st series [0th ⇒ 6th ⇒ 5th ⇒ 4th ⇒ 3rd]   2nd series [0th ⇒ 3rd ⇒ 4th ⇒ 5th ⇒ 6th]

4. The first output value $x_k^a$ of the cue neuron in $a$ −Cue Ball is calculated by inputting the memorized image pattern into the corresponding Recall Net.

5. Using $x_k^a$ obtained in step 4, the intermediate output $q_l^b$ of the cue neuron in the next $b$ −Cue Ball is calculated.

6. Using $q_l^b$ and $x_k^a$ obtained in step 5, the connection weights of Eqs. (16) and (17) are learned.

7. Save the learned values of $u_{kl}^b$ to a file.

8. Return to step 2 and continue processing each series set in step 3.

   ※ Note that the learning value $\theta$ is set to 100 for 1st series and to 110 for 2nd series.

9. Return to step 1 and repeat steps 1 to 8 above with the Cue Ball combination order determined by cmb=2.

In the following, we describe the algorithm for the recall process. In the recall process, a trained QR code pattern image is first presented to the Recall Net corresponding to the $a$ −Cue Ball. At this time, the pattern image data is input to all cue neurons of the $a$ −Cue Ball, but only the $k$-th cue neuron that has trained the presented pattern generates an output ($x_k^a = 1.0$). This output value is input to all learned cue neurons of $b$ −cue ball to examine their outputs. Now find the $q_l^b$ of the neuron that produces the maximum output and use a threshold function to calculate $x_l^b$ (which have a value of 1.0). If this output is input to all learned recall neurons of the corresponding Recall Net, the corresponding learned QR-code pattern image will be displayed. After this, we use $x_l^b$ (= 1.0) to check the output of all the cue neurons in the next $c$ −Cue Ball and do the same process as above. This process is repeated until the final Cue Ball is reached. A summary of these recall algorithms is provided below.

[Algorithm 4 - Associative recall between learned cue neurons for different Cue Balls]

1. Specify the group to be presented (cmb=0,1) (see Algorithm 3)

2. The QR code pattern images learned by the cue neurons in $a$ −Cue Ball are presented to the corresponding Recall Net.

   ※ The presented pattern images are in the order of the cue neuron numbers specified in step 3 of Algorithm 3.

3. The obtained output value "$x_k^a (= 1.0)$" is input to all cue neurons in $b$ −Cue Ball, and the output

value "$q_l^b$" is calculated using eq. (14).

4. The largest value obtained, $q_l^b$, is used to calculate "$x_k^b (= 1.0)$" using a threshold function.
   ※ When the output value "$x_k^b (= 1.0)$" is used, the memorized QR code image pattern will be displayed in Recall Net.

5. This output value "$x_k^b = 1.0$" is input to all cue neurons in c-Cue Ball, and the same process as step 3 to step 4 is performed.

6. Repeat these steps until $e-$Cue Ball is reached.

7. Once the "cmb = 0" process is complete, return to step 2 and perform the "cmb = 1" process in the same way.

In the following, we conduct simulations using these algorithms to verify whether learning and recall proceed as expected. Figure 3 shows the output values $q_i^{attr}$ of all cue neurons (0th to 6th) for each Cue Ball (Color, Shape, Volume, Spectacular View, and Constellation) after learning $w_{ji}^{attr}$ and $v_{ij}^{attr}$ according to Algorithms 1 and 2.

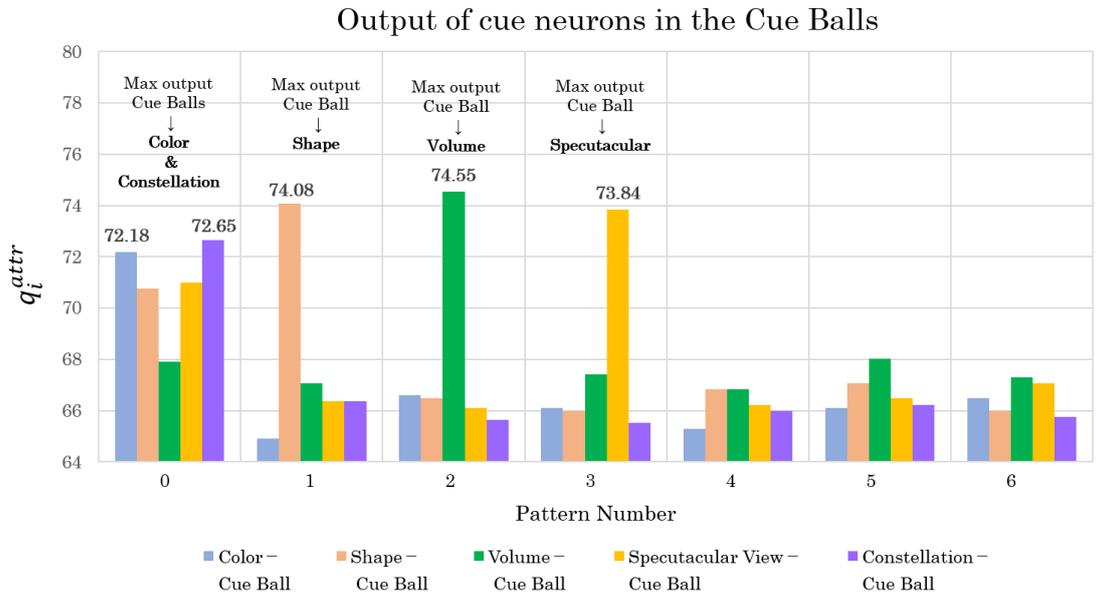

Figure 3: $q_i^{attr}$ of the cue neuron for each Cue Ball corresponding to the learning pattern number. Each Cue Ball, we examined the output values of the 0th to 6th neurons. In the Color−Cue Ball, the QR-code image that the 0th neuron has been learning has presented to the Recall Net, and the output values of the 0th to 6th neurons have examined. Similarly, we have presented to the Recall Net the learned images to the 1st neuron for Shape−Cue Ball, the 2nd neuron for Volume−Cue Ball, and the 3rd neuron for Spectacular View−Cue Ball, and have examined the output values of each cue neuron. In addition, for use in later simulation processing, in Constellation−Cue Ball, the image learned by the 0th neuron, rather than the 4th neuron, is presented to the Recall Net, and the results of calculating the output values of each cue neuron are shown. The final output value of the neuron that outputs the largest "$q_i^{attr}$" in each image pattern is "$x_i^{attr} = 1.0$."

For example, in the Color-Cue Ball, the QR-code image that the 0th neuron has been learning has presented to the Recall Net, and the output values of the 0th to 6th neurons have examined. In this case, the intermediate output value "$q_0^{Color}$" of the 0th neuron learned has been the largest, and all other values have been smaller than this. Similarly, we presented the pattern image learned by the first cue neuron for the Shape-Cue Ball, the pattern image learned by the second cue neuron for the Volume-Cue Ball, and the pattern image learned by the third cue neuron for the Spectacular View-Cue Ball to the corresponding Recall Net, and examined the output values of all cue neurons for each Cue Ball. The results are shown in Fig. 3. For Constellation-Cue Ball, normally, this procedure would involve using the QR code pattern image trained with the fourth cue neuron, but since the results obtained here will be used in a simulation of the subsequent Recall process algorithm, we designated the zeroth cue neuron as the trained pattern image and examined the output values of each cue neuron. The $q_i^{attr}$ values when an image different from the pattern image specified above (a cue neuron with a different number) was presented for each cue ball are also shown. Note that the data for Figure 3 is shown in Table 2.

| | $q_i^{attr}$ | | | | |
|---|---|---|---|---|---|
| cue neuron No. | Color Cue Ball | Shape Cue Ball | Volume Cue Ball | Spectacular View Cue Ball | Constellation Cue Ball |
| 0 | **72.18** | 70.75 | 67.90 | 70.99 | **72.65** |
| 1 | 64.92 | **74.08** | 67.06 | 66.35 | 66.35 |
| 2 | 66.59 | 66.47 | **74.55** | 66.11 | 65.64 |
| 3 | 66.11 | 65.99 | 67.42 | **73.84** | 65.52 |
| 4 | 65.28 | 66.83 | 66.83 | 66.23 | 65.99 |
| 5 | 66.11 | 67.06 | 68.01 | 66.47 | 66.23 |
| 6 | 66.47 | 65.99 | 67.30 | 67.06 | 65.76 |

Table.2: Output values of all cue neurons in each Cue Ball in Figure.2. The output value $q_i^a$ of each cue ball's cue neuron is shown. Note that the dark red and bold text indicates the maximum value.

As shown in Figure 3, in the Color–Cue Ball, the pre-learned 0 –th cue neuron has output the maximum output value $q_{i=0}^{Color} = 72.18$. When the threshold function is applied to this $q_{i=0}^{Color}$, the result is "$x_{i=0}^{Color} = 1.0$", which means that the first pattern is recalled as expected, confirming that learning is complete. In the Shape-Cue Ball, the 1st learned cue neuron outputs the maximum output value, "$q_{i=1}^{Shape} = 74.08$", resulting in "$x_{i=1}^{Shape} = 1.0$", which recalls the expected pattern. Similarly, in the Volume–Cue Ball, the 2nd learned cue neuron outputs the maximum output value "$q_{i=2}^{Volume} = 74.55$", resulting in "$x_{i=2}^{Volume} = 1.0$", and in the Spectacular View–Cue Ball -Cue Ball, the 3rd learned cue neuron outputs the maximum output value "$q_{i=3}^{Specutacular\ View} = 73.84$", resulting in "$x_{i=3}^{Spectacular\ View} = 1.0$", which recalls the expected pattern. In Constellation−Cue Ball, the 0th learned cue neuron is specified, and this cue neuron outputs the maximum output value, "$q_{i=0}^{Constellation} = 72.65$", resulting in "$x_{i=0}^{Constellation} = 1.0$", which recalls the first pattern as expected. Based on these simulation results, the maximum value of $q_i^{attr}$ across the five types of learned cue neurons exceeds 72.0; therefore, the threshold used in each algorithm has set to "$D = 72.0$." As a side note, when the 4th learned cue neuron is specified for the Constellation-Cue Ball, it produces an output of "$q_{i=4}^{Constellation} = 72.89$", resulting in "$x_{i=4}^{Constellation} = 1.0$."

Finally, let's run Recall Algorithm 4. As shown in Algorithm 3, two learning series are performed for each groups (cms=1,2). For cmb = 0, it is assumed that, in response to the output of the 0th cue neuron of the Color-Cue Ball, the 1st and 4th cue neurons among all cue neurons of the Shape-Cue Ball produce outputs. Following this, the 2nd and 3rd cue neurons of the Volume-Cue Ball are assumed to be recalled, followed by the 3rd and 2nd cue neurons of the Spectacular View-Cue Ball, and finally the 4th and 1st cue neurons of the Constellation-Cue Ball, resulting in a chain of recall events. A similar procedure is applied to the second series. These procedures are summarized in Table 3.

| cmb=0 | | The Starting Cue Ball : *Color* | | | | |
|---|---|---|---|---|---|---|
| Learning series | Cue Ball | *Color* | *Shape* | *Volume* | *Spectacular View* | *Constellation* |
| Serise1 | Learning cue neurons | 0th ⇒ | 1st ⇒ | 2nd ⇒ | 3rd ⇒ | 4th |
| Serise2 | | | 4th ⇒ | 3rd ⇒ | 2nd ⇒ | 1st |

| cmb=1 | | The Starting Cue Ball : *Constellations* | | | | |
|---|---|---|---|---|---|---|
| Learning series | Cue Ball | *Constellation* | *Spectacular View* | *Volume* | *Shape* | *Color* |
| Serise1 | Learning cue neurons | 0th ⇒ | 6th ⇒ | 5th ⇒ | 4th ⇒ | 3rd |
| Serise2 | | | 3rd ⇒ | 4th ⇒ | 5th ⇒ | 6th |

Table.3: Two groups (cmb=0, 1) of recall process and each series. For both cmb = 0 and cmb = 1, the outputs of the learned cue neurons within each of the five Cue Balls are transmitted sequentially. For example, if cmb = 0, a learned QR code pattern image is presented to the recall net corresponding to the color cue ball. In this case, the 0th cue neuron, which is training the presented image, generates an output. Then, based on the output of this cue neuron, the first and fourth cue neurons of Shape-Cue Ball output. Repeat the same process until you reach the Constellation−Cue Ball. Execute the same for "cmb=1."

The cue ball groups to be chained are cmb = 0 and cmb = 1, and these two groups are executed in sequence. For the cmb = 0 group, the learned QR code pattern image is first presented to the Color−Cue Ball, and the outputs of all cue neurons in this Cue Ball are examined. At this case, the learned 0th cue neuron has an output, with an intermediate output of "$q_{i=0}^{Color} = 72.18$" and a final output of "$x_{i=0}^{Color} = 1.0$." This output causes outputs in the 1st and 4th learned cue neurons of Shape−Cue Ball. Furthermore, if we use this output and apply the same procedure as above, the 2nd and 3rd cue neurons of the Volume−Cue Ball, the 3rd and 2nd cue neurons of the Spectacular View−Cue Ball, and the 4th and 1st cue neurons of the Constellation−Cue Ball output in that order. The same procedure is subsequently applied to the cmb = 1 group. First, the learned QR code pattern image is presented to the Constellation−Cue Ball, and the outputs of all cue neurons in this Cue Ball are examined. At this stage, the learned 0th cue neuron has an output, with an intermediate output of "$q_{i=0}^{Constellation} = 72.65$" and a final output of "$x_{i=0}^{Constellation} = 1.0$." In this case, the output of the 0th cue neuron, learned in the same way as "cmb = 0" is the intermediate output "$q_{i=0}^{Constellation} = 72.65$" and the final output "$x_{i=0}^{Constellation} = 1.0$." Furthermore, if we use this output and apply the same procedure as above, the 6th and 3r cue neurons of the Spectacular View−Cue Ball, the 5th and 4th cue neurons of the Volume−Cue Ball, 4th and 5th cue neurons of the Shape-Cue Ball and the 2nd and 6th cue neurons of the Color−Cue Ball output in that order.

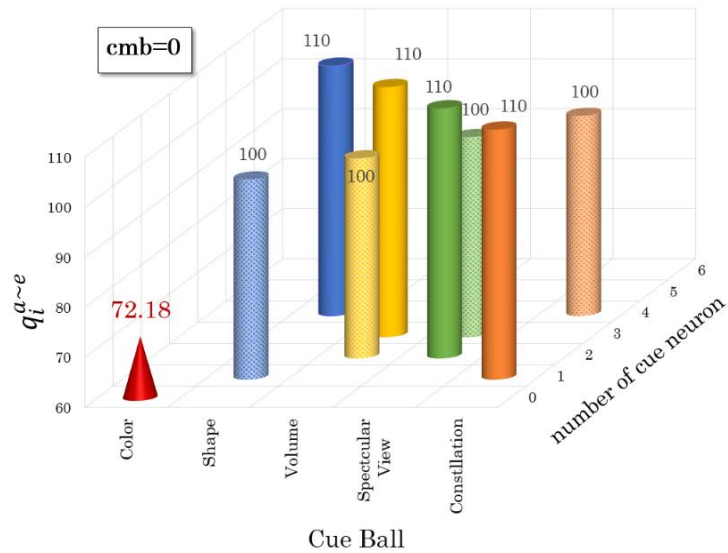

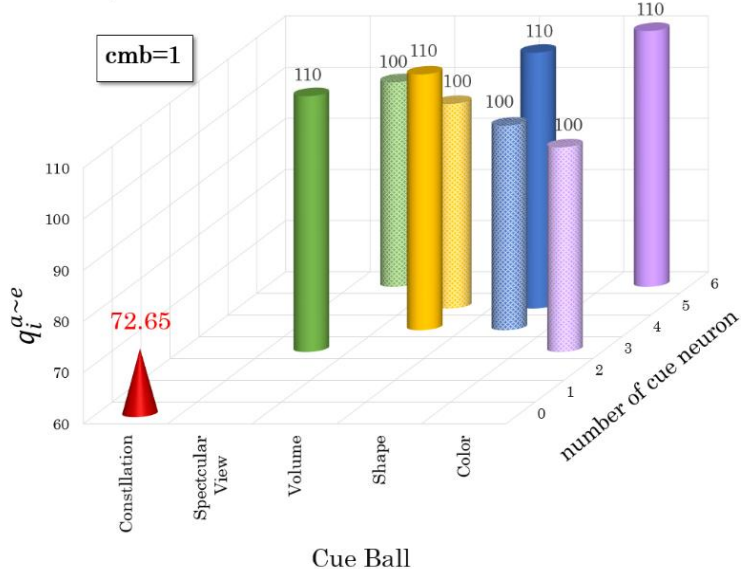

Figure.4: The chain output of the leaned cue neurons for each Cue Ball. The top row is the "cmb=0" group, and the bottom row is the cmb=1 group. When "cmb = 0", when a QR code pattern image corresponding to a colored Cue Ball is presented to the recall net, the cue neuron learning this image has outputted "$q_{i=0}^{Color} = 72.18$." This output causes the Shape−Cue Ball's learned 1st and 4th cue neurons to be output. Using the output values of these two cue neurons, we can confirm the outputs of the 2nd and 3rd cue neurons learned by Volume-Cue Ball. Similarly, the Spectacular-Cue Ball outputs the learned cue neurons 2nd and 3rd, followed by the Constellation-Cue Ball's learned cue neurons 1st and 4th. The results of a similar treatment with "cmb=1" are also shown. Also, the output values 100 and 110 at the top of each graph correspond to learning coefficients $\theta$ of 100 and 110, respectively.

The simulation results show that when a trained QR code pattern image is presented to the cue neuron of the first Cue Ball, subsequent learned images are recalled in a chain reaction. In other words, (1 + 4 × 2) cue neurons respond, recalling nine different images. The above simulation process has two types, "cmb = 0, 1", so ultimately "9 types of images × 2 groups = 18 types of images" will be displayed. Also, all cue neurons that showed large outputs were cue neurons learned in this simulation. Furthermore, for each cmb, two learning sequences were performed, and accordingly the learning coefficient $\theta$ has been set separately to 100 and 110. When the learning coefficient θ has been 100, all learned cue neurons that have been associatively recalled except for the starting cue neuron have been 100, and when θ has been 110, all have been 110. In addition, the outputs of all other unlearned cue neurons have been 0. This has confirmed that all learning and recall processes are functioning as expected.

4. Conclusions and Discussions

In this paper, we introduced a neural network model of associative memory [3-9] that can learn attributes as images and continuously recall multiple learned memories. Basically, this model has extended the previous attribute-specific associative memory neural network model [1] by increasing the number of attributes considered and allowing for further connections between attributes. Therefore, the basic configuration of the model is the conventional CB-RN system [1] [10]. The number of linked attributes considered here was set to 5 based on reports that humans can associate approximately five memories at a time [31–34]. To represent attributes, the name of the attribute element has been treated as a QR-code pattern image. The provided attribute types consist of five systems: the C.CB-RN system for processing color (Color), the S.CB-RN for shape (Shape), the V.CB-RN for volume (Volume), the SV.CB-RN for processing the names of spectacular views (Spectacular View), and the CN.CB-RN for processing constellation names (Constellation). Each attribute is composed of seven attribute elements. Although there is no particular limitation on increasing the number of attribute elements or even the number of attributes, we consider the present configuration to be sufficient for an experimental model. In this model, representing attribute element names and related information as QR-code pattern images offers the advantage of dramatically expanding the range of possible applications. Furthermore, subtle tastes, odors, and other sensory perceptions can also be represented within this framework. In the simulation, the neurons in each Cue Ball and Recall Net have been learned using QR-code pattern images (digital data), and moreover further learning has been performed between cue neurons of different Cue Balls. As a result, when a QR-code pattern image was presented to the recall neuron group of one Recall Net, the learned cue neurons responded, and it was confirmed that the learned cue neurons in different Cue Balls then responded one after another. As a result, the pattern images of the learned QR-codes are displayed on the relevant Recall Nets. In addition, for each cmb group, two learning sequences were executed with different learning coefficients $\theta$ (100 and 110), resulting in differences in the output values of the learned cue neurons. When the learning coefficient $\theta$ was set to 100, the output values of all associatively recalled learned cue neurons—excluding the origin cue neuron—were 100; when $\theta$ was set to 110, all such output values were 110. It is thought that this difference in output value can be used to determine the strength of recollection.

In general, it is believed that human associative memory operates such that recalling one item triggers the successive recall of related memories. We consider the brain repeatedly performs learning based on memory data accumulated in this manner. Also, memory recall appears to be closely related to imagery. From this perspective, research into the mechanisms of memory, especially image recall, is considered to be a very important field. In the associative memory model based on neural networks presented in this paper, presenting a learned image enables the successive recall of related images in a manner analogous to human memory processing [3, 6–9, 28–29]. Therefore, we believe that the model presented here captures some aspects of the mechanisms of human memory and recall. In this paper, we standardized the size (pixel dimensions) of QR code pattern images, but in reality, image sizes are generally non-uniform. In future research, we would like to consider how to deal with such cases.

## Appendix

All QR-code pattern images used in this paper are shown below. The numbers at the top of the image (such as 0th) are the numbers of the corresponding cue neurons. The file name (e.g., "red") at the bottom of each pattern, excluding the extension, is the string saved as the QR-code image pattern.

| Attributes Members in Color－Cue Ball | | | | | | |
|---|---|---|---|---|---|---|
| 0th | 1st | 2nd | 3rd | 4th | 5th | 6th |
| 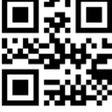 | 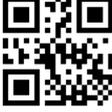 | 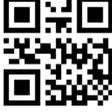 | 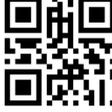 | 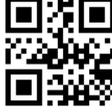 | 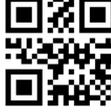 | 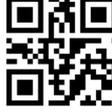 |
| red.bmp | orangae.bmp | yellow.bmp | green.bmp | blue.bmp | indigo.bmp | purple.bm |
| Attributes Members in Shape－Cue Ball | | | | | | |
| 0th | 1st | 2nd | 3rd | 4th | 5th | 6th |
| 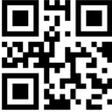 | 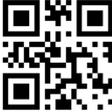 | 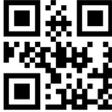 | 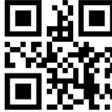 | 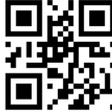 | 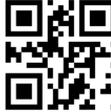 | 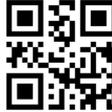 |
| square.bmp | circle.bmp | oval.bmp | Rectangle.bmp | Trapezoid.bmp | Triangle.bmp | rhombus.bm |
| Attributes Members in Volume－Cue Ball | | | | | | |
| 0th | 1st | 2nd | 3rd | 4th | 5th | 6th |
| 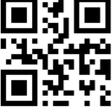 | 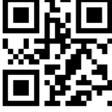 | 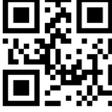 | 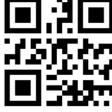 | 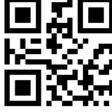 | 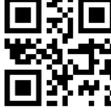 | 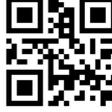 |
| extra-large.bmp | large.bmp | medium.bmp | small-medium.bmp | small.bmp | extra-small.bmp | mini.bm |
| Attributes Members in Spectacular View－Cue Ball | | | | | | |
| 0th | 1st | 2nd | 3rd | 4th | 5th | 6th |
| 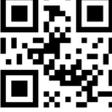 | 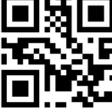 | 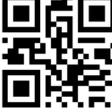 | 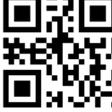 | 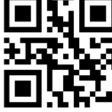 | 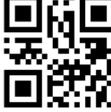 | 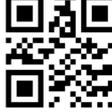 |
| Iguazu.bmp | Maunakea.bmp | Milford Sound.bmp | Monument VY.bmp | Rockies.bmp | Tekapo.bmp | Yellowknife.bmp |
| Attributes Members in Constellation－Cue Ball | | | | | | |
| 0th | 1st | 2nd | 3rd | 4th | 5th | 6th |
| 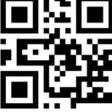 | 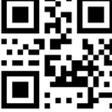 | 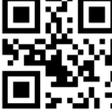 | 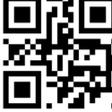 | 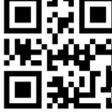 | 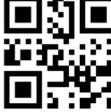 | 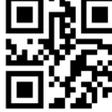 |
| Andromeda.bmp | Aquarius.bmp | Cassiopeia.bmp | Centaurus.bmp | Cygnus.bmp | Orion.bmp | Perseus.bmp |